\documentclass{article} % For LaTeX2e
\usepackage{iclr2025_conference,times}
\usepackage{algorithm}
\usepackage{algorithmicx}
\usepackage{algpseudocode}
\usepackage{booktabs,csquotes,verbatim}    
\usepackage{needspace}
\usepackage{amsmath,amsfonts,bm}

\def\eqref#1{equation~\ref{#1}}
\def\1{\bm{1}}

\DeclareMathAlphabet{\mathsfit}{\encodingdefault}{\sfdefault}{m}{sl}
\SetMathAlphabet{\mathsfit}{bold}{\encodingdefault}{\sfdefault}{bx}{n}

\DeclareMathOperator*{\argmax}{arg\,max}

\usepackage{hyperref}
\usepackage{url}
\usepackage{placeins}
\usepackage{caption}
\usepackage{float}
\restylefloat{algorithm}

\usepackage{graphicx}

\iclrfinalcopy

\title{Evolutionary Prompt Optimization Discovers Emergent Multimodal Reasoning Strategies in Vision-Language Models}

\author{%
  Sid Bharthulwar\thanks{Equal contribution. Correspondence to sidbharthulwar@college.harvard.edu. } \quad
  John Rho\footnotemark[1] \quad
  Katrina Brown\footnotemark[1] \\
  Harvard College \\
}

\newboolean{comments}
\setboolean{comments}{true}%\showComments}
\ifthenelse{\boolean{comments}} {
	\newcommand{\sid}[1]{\textcolor{red}{(Sid: #1)}}
	\newcommand{\john}[1]{\textcolor{blue}{(John: #1)}}
    \newcommand{\katrina}[1]{\textcolor{purple}{(katrina: #1)}}
} {
	\newcommand{\sid}[1]{}
	\newcommand{\john}[1]{}
    \newcommand{\katrina}[1]{}
}

\begin{document}

\maketitle

\begin{abstract}
    % We present an evolutionary framework for optimizing prompts in vision-language models that achieves significant improvements in multimodal reasoning tasks without requiring model retraining. Our approach introduces three key innovations: (1) a hierarchical evolutionary system that operates across task, mutation, and meta-mutation prompt spaces; (2) an LLM-based fitness function tuned for vision that combines task performance with prompt quality metrics; and (3) a novel mechanism for emergent tool synthesis that allows evolved prompts to decompose complex visual tasks into iterative reasoning steps. Evaluations on multiple benchmarks, including MathVista, M3CoT, and GeoBench-VLM, demonstrate substantial improvements over baseline approaches, with particularly strong gains in tasks requiring complex spatial and physical reasoning. For instance, our method improves performance on visual question-answering tasks by up to $\approx 11$ percentage points over the base model and achieves a 49\% relative improvement in damage assessment tasks. Most notably, these improvements emerge through purely inference-time optimization, suggesting that sophisticated reasoning strategies can be unlocked in vision-language models without expensive retraining procedures. Our findings indicate that evolutionary prompt optimization represents a promising direction for enhancing multimodal AI systems while maintaining computational efficiency and interpretability.%
    We present a framework for optimizing prompts in vision-language models to elicit multimodal reasoning without model retraining. Using an evolutionary algorithm to guide prompt updates downstream of visual tasks, our approach improves upon baseline prompt-updating algorithms, which lack evolution-style "survival of the fittest" iteration. Crucially, we find this approach enables the language model to independently discover progressive problem-solving techniques across several evolution generations. For example, the model reasons that to "break down" visually complex spatial tasks, making a tool call to a Python interpreter to perform tasks (such as cropping, image segmentation, or saturation changes) would improve performance significantly. Our experimentation shows that explicitly evoking this "tool calling" call, via system-level XML $...\texttt{<tool>} ... \texttt{</tool>}...$ tags, can effectively flag Python interpreter access for the same language model to generate relevant programs, generating advanced multimodal functionality. This functionality can be crystallized into a system-level prompt that induces improved performance at inference time, and our experimentation suggests up to $\approx 50\%$ relative improvement across select visual tasks. Downstream performance is trained and evaluated across subtasks from MathVista, M3CoT, and GeoBench-VLM datasets. Importantly, our approach shows that evolutionary prompt optimization guides language models towards self-reasoning discoveries, which result in improved zero-shot generalization across tasks. 
\end{abstract}

\section{Introduction}
Vision-language models (VLMs) have made noteworthy progress in simultaneously understanding and generating text and images, with recent work extending large language models (LLMs) to operate effectively across multiple modalities \citep{chameleon,transfusion}. Despite these advances, current foundation models typically treat visual and textual reasoning as loosely connected processes \citep{zhang2024improvevisionlanguagemodel,chen2024measuringimprovingchainofthoughtreasoning}, often underutilizing their full potential for \emph{native} multimodal reasoning.

In the text-only realm, \emph{chain-of-thought} (CoT) prompting and its variants like zero-shot prompting and self-consistency decoding have emerged as powerful paradigms for enabling systematic reasoning \citep{wei2023chainofthoughtpromptingelicitsreasoning,zero_shot_cot_ea_2024,wang2023selfconsistencyimproveschainthought}. While recent work has adapted CoT to vision-language models through rationale distillation or latent space training \citep{zhang2024improvevisionlanguagemodel,hao2024traininglargelanguagemodels}, these approaches often require expensive retraining or specialized datasets.

A complementary research direction seeks to enhance reasoning purely at inference time through prompt optimization. Methods like Promptbreeder \citep{fernando2023promptbreederselfreferentialselfimprovementprompt} and evolutionary algorithms \citep{guo2024connecting} have shown promise in text-only scenarios, but their application to vision-language tasks remains understudied. The challenge of discovering effective \emph{multimodal} reasoning strategies through purely inference-time evolution presents unique opportunities and challenges.

In this paper, we propose a novel \emph{evolutionary prompt optimization} framework tailored for vision-language tasks. Our method operates on standard VLMs without any model retraining, evolving task-specific prompts that coordinate visual and linguistic processing. We evaluate our approach on challenging multimodal benchmarks requiring commonsense physics, complex counting, and mathematical reasoning. We demonstrate that evolutionary search can discover sophisticated emergent behaviors such as recursive tool usage, hierarchical image partitioning, and dynamic-programming–like counting entirely through prompt mutations. Critically, the performance gains achieved by our methods require as few as 20 labeled examples per subtask, and generalize robustly to unseen data. 

%Our framework yields significant improvements over baselines, including a 14.5 percentage point gain on physical commonsense tasks and nearly 50\% relative improvement in geospatial counting tasks. 
%Importantly, our experimental gains require as few as 20 labeled examples per subtask, as shown in Figure~\ref{gen-chart}, suggesting high generalization potential.

Our contributions are the following: \textbf{(1)}~ Evolutionary search in natural language prompts can uncover natively multimodal reasoning strategies in vision-language models. \textbf{(2)}~Purely inference-time prompt evolution can discover sophisticated behaviors that mirror algorithmic reasoning. \textbf{(3)}~Extensive evaluations across multiple benchmarks demonstrate significant improvements and shed light on the synergy between textual instructions and visual inputs.

\section{Related Works}

\subsection{Chain-of-Thought Prompting in Language and Vision-Language Models}

Chain-of-Thought (CoT) prompting has emerged as a powerful paradigm for enabling Large Language Models (LLMs) to solve complex tasks by generating intermediate reasoning steps before arriving at a final answer~\citep{zhang2022automaticchainthoughtprompting}. The success of CoT prompting has extended to Vision-Language Models (VLMs) in recent work. \citet{zhang2024improvevisionlanguagemodel} highlight that relying solely on brief annotations constrains the depth of multimodal reasoning. By distilling more comprehensive rationales from GPT-4 and incorporating reinforcement learning signals, they significantly enhance the interpretability and robustness of VLM outputs. Similarly, \citet{chen2024measuringimprovingchainofthoughtreasoning} examine the consistency of VLM reasoning and propose methods for systematically quantifying and improving step-by-step visual grounding. While these advances have led to more transparent VLM behaviors, they frequently rely on large-scale datasets or specific fine-tuning stages for reliable CoT generation.
% TODO: add a bit more about scaling time inference compute, maybe mentioning o1 and r1

\subsection{Unified Multimodal Reasoning in Vision-Language Models}

Recent studies have sought to broaden the scope of modern foundation models from reasoning purely over text tokens to fully multimodal reasoning. Chameleon \citep{chameleon} and Transfusion (\citep{transfusion} enable multimodal reasoning by enabling transformers to natively generate mixed-modal tokens. However, no works have combined these omnimodal models with advanced methods in reasoning, such as scaling inference-time compute or automatic prompt optimization. With the recent successes in inference-time compute, such as OpenAI's O1 and DeepSeek's R1, this is a highly motivated problem of study \citep{deepseekai2025deepseekr1incentivizingreasoningcapability}. Additionally, \cite{hao2024traininglargelanguagemodels} shift from utilizing purely textual chains of thought to continuous latent spaces, and Zhang et al and Chen et al leverage massive curated rationales from teacher models to bridge the modality gap in reasoning. Although these methods show promise, they often require significant model re-training or specialized data collection. Our approach diverges from these by focusing on a purely \emph{inference-time} adaptation that doesn't require expensive pretraining or posttraining procedures.

\subsection{Evolutionary Algorithms for Prompt Optimization}

Parallel to these developments in vision-language reasoning, evolutionary algorithms have been increasingly employed to optimize prompts in LLMs. \citet{guo2024connecting} introduce EvoPrompt, demonstrating that even discrete natural language prompts can be systematically evolved to enhance downstream task accuracy. \citet{zero_shot_cot_ea_2024} similarly leverage evolutionary strategies to iteratively refine zero-shot CoT prompts, highlighting that diversity in prompt mutations can mitigate potential blind spots in static prompts. Moving a step further, \citet{fernando2023promptbreederselfreferentialselfimprovementprompt} propose Promptbreeder, which uses self-referential, self-improving prompt mutation to outperform standard CoT methods on arithmetic and commonsense benchmarks.

Beyond evolutionary approaches, there has been extensive recent work on automatic or optimized prompting that does \emph{not} rely on EAs. Methods such as RLPrompt \citep{deng-etal-2022-rlprompt}, adversarial in-context learning \citep{do2024promptoptimizationadversarialincontext}, InstructZero \citep{chen2023instructzeroefficientinstructionoptimization}, Instinct \citep{lin2024useinstinctinstructionoptimization}, and gradient-based or search-based prompt optimization \citep{pryzant2023automaticpromptoptimizationgradient, rubin2022learningretrievepromptsincontext, sun2023autohintautomaticpromptoptimization,  wan2024teachbettersmarterinstructions, zhou2023largelanguagemodelshumanlevel, yang2024largelanguagemodelsoptimizers} collectively demonstrate the general interest in systematically refining prompt instructions without exhaustive retraining of large language models. However, these strategies remain mostly text-centric and do not directly address complex multimodal tasks as in vision-language scenarios.

Despite the demonstrated success of these frameworks in purely text-based environments, the vision-language domain brings additional challenges, such as object localization, visual attribute recognition, and scene context understanding. Additionally, these works fail to elicit truly novel reasoning paradigms from evolutionary search. Thus, bridging evolutionary search techniques with multimodal CoT remains an open research question. Our work takes a step in this direction by proposing a purely \emph{inference-time} evolutionary approach that integrates visual and textual prompts for improved multimodal reasoning.

\section{Methodology}

\subsection{Problem Formulation}
We address the challenge of evolving task-specific prompts for vision-language models (VLMs) applied to downstream multimodal reasoning tasks through an evolutionary prompt optimization framework. Given a high-level multimodal cognitive task such as counting or classification, we seek to discover optimal system prompts that enhance model performance across diverse problem instances within that given task.

Let $Q$ represent the complete set of question instances for a specific task. For any question $q \in Q$, we define a system prompt $p$ from the prompt space $\mathcal{P}$ that is prepended to the question. The concatenation operation $\oplus: \mathcal{P} \times Q \rightarrow \mathcal{S}$ maps a prompt-question pair to the final input string space $\mathcal{S}$. Our objective function is then: $ p^* = \argmax_{p \in \mathcal{P}} \mathbb{E}_{q \in Q}[\text{Score}(p \oplus q)]$ where $\text{Score}: \mathcal{S} \rightarrow [0, 100]$ evaluates the quality of the LLM's response. The score normalization to $[0, 100]$ enables consistent comparison across different task types, with task-specific metrics (e.g., accuracy for classification, precision for counting) mapped to this standardized range.

We partition $Q$ into training and test sets $A$ and $B$, respectively. Our method relies on a fundamental assumption about the objective function's behavior across these sets:
\begin{equation}
    \argmax_{p \in \mathcal{P}} \mathbb{E}_{q \in A}[\text{Score}(p \oplus q)] \approx \argmax_{p \in \mathcal{P}} \mathbb{E}_{q \in B}[\text{Score}(p \oplus q)]
\end{equation}
This assumption allows us to optimize prompts on the training set with the expectation that they will generalize effectively to unseen test instances. For practical utility of our technique on downstream tasks, we operate in the regime where $|B| >> |A|$, with the assumption that the evolutionary framework can generalize past the train set and find the globally optimal task-specific prompt $p^*$. The practical validity of this assumption is demonstrated empirically in Section 5. The small training set serves as a "few-shot training set" for evolution, so that the large test set is indeed the real target.

\subsection{Evolutionary Algorithm Design}
Our evolutionary framework operates across three hierarchical spaces: the task prompt space $\mathcal{P}$, mutation prompt space $\mathcal{M}$, and meta-mutation prompt space $\mathcal{H}$. This hierarchical structure enables both direct optimization of task prompts and meta-learning of effective mutation strategies.

\subsubsection{Population Evolution}

\begin{algorithm}[H]
    \caption{Binary Tournament Evolution}
    \begin{algorithmic}[1]
        \For{generation $g = 1$ to $G$}
        \State Sample prompts $p_1, p_2 \sim \mathcal{P}_g$ without replacement
        \State $p_w \leftarrow \arg \max_{p \in \{p_1, p_2\}} \text{Fitness}(p)$
        \State $p_l \leftarrow \arg \min_{p \in \{p_1, p_2\}} \text{Fitness}(p)$
        \State $m \sim \mathcal{M}$ \Comment{Sample mutation prompt}
        \State $p'_w \leftarrow \text{LLM}(m \oplus p_w)$ \Comment{Mutate winner}
        \State $\mathcal{P}_{g+1} \leftarrow (\mathcal{P}_g \setminus \{p_l\}) \cup \{p'_w\}$
        \EndFor
    \end{algorithmic}
\end{algorithm}

Our framework uses a binary tournament genetic algorithm to evolve task prompts. %At each generation $g$, we maintain a population $\mathcal{P}_g$ that evolves through selection and mutation. The algorithm samples two prompts without replacement, evaluates them using fitness function $F(p)$ (which considers both task performance and prompt quality), and designates a winner $p_w$ and loser $p_l$. Mutation focuses exclusively on the winner, using a sampled mutation prompt $m \in \mathcal{M}$ to generate $p'_w = \text{LLM}(m \oplus p_w)$. This leverages the LLM's language understanding to create meaningful variations.%
The population evolves by replacing $p_l$ with $p'_w$, maintaining size while improving fitness over $G$ generations. The binary tournament balances exploration and exploitation, reduces computational overhead, and creates selection pressure towards better solutions. Mutation combines structured guidance from $\mathcal{M}$ with the LLM's flexibility, enabling discovery of effective prompts that manual or fully automated approaches might miss.

\subsubsection{Mutation Operators}
Our framework employs a hierarchical system of mutation operators that combines both zero-order and first-order optimization strategies. The fundamental mutation process occurs in the prompt space $\mathcal{P}$, where each prompt $p \in \mathcal{P}$ represents a strategy for solving a given task. These mutations are guided by prompts from the mutation space $\mathcal{M}$ and hyper-mutation space $\mathcal{H}$.

We employ both first-order and zero-order prompt optimization techniques. For first-order optimization, we generate a new task prompt by applying the mutation prompt to the current prompt:
\begin{equation}
    p' = \mu_1(p, m) = \mathcal{L}(m \oplus p)
\end{equation}
where $\mathcal{L}$ represents the language model's text generation function and $\oplus$ denotes concatenation.

For zero-order optimization, we generate a new task prompt independently by concatenating the problem description $D$ with a hint-generation template:
\begin{equation}
    p'' = \mu_0(D) = \mathcal{L}(\text{"A list of 100 hints:"} \oplus D)
\end{equation}
This allows for the generation of novel task prompts that are closely tied to the original problem description, providing diversity in the evolutionary process.

The mutation prompt $m$ itself evolves through both first-order and zero-order hyper-mutation operators. The first-order hyper-mutation operator is defined as:
\begin{equation}
    m' = \nu_1(m, h) = \mathcal{L}(h \oplus m)
\end{equation}
where $h \in \mathcal{H}$ is a hyper-mutation prompt.

The zero-order hyper-mutation operator generates new mutation prompts by combining the problem description with a hint-generation template, simliar to the zero-order mutation operator.
\begin{equation}
    m'' = \nu_0(D, t) = \text{"A list of 100 hints:"} \oplus D
\end{equation}

We adapt this paradigm of zero and first-order prompt optimization from Promptbreeder, and find that it generalizes well across vision-language tasks when initial prompt populations are vision-specific. This hierarchical system allows for both direct optimization of task prompts and adaptation of mutation strategies, while maintaining simplicity and interpretability in the evolutionary process. The combination of zero-order and first-order operators ensures both exploration of new ideas and refinement of existing solutions.

\subsection{Fitness Evaluation}
The fitness function $F: \mathcal{P} \rightarrow \mathbb{R}$ evaluates task prompts through a weighted combination of task performance and prompt quality metrics, defined as $F(p) = (1-\lambda)F_{\text{task}}(p) + \lambda F_{\text{aux}}(p)$. The task fitness component $F_{\text{task}}(p) = \frac{1}{k}\sum_{q \in C} \text{Score}(p \oplus q)$ measures empirical performance on a stochastically determined minibatch $C \subset A$ of size $k$, where $\text{Score}: \mathcal{S} \rightarrow [0, 100]$ quantifies the quality of the LLM's response to task instance $q$ when using prompt $p$. The score function is task-specific--for example, in the case of a counting task, the score function is the percentage of correct answers.

The auxiliary fitness component $F_{\text{aux}}(p) = \mathcal{L}_{\text{critic}}(\text{critique} \oplus p)$ employs an LLM-based critique system that evaluates the sensibility and adherence of task prompts to their intended goals. This critique system acts as a regularizer for the evolutionary search process, steering the optimization towards prompts that are not only effective but also semantically meaningful and aligned with the task objectives. This aligns with previous literature that leverage LLMs' expansive knowledge base for optimization tasks, even in settings of sparse reward \citep{yang2024largelanguagemodelsoptimizers}. We find that critique prompts that emphasize the coherence, explicitness, and adherence of mutated task prompts to standard formatting perform the best, effectively guiding the evolutionary search process. By incorporating this linguistic prior, we significantly improve sample efficiency, as demonstrated empirically in Section 5, while maintaining the discovery of high-performing prompts.

The weighting coefficient $\lambda$ balances the trade-off between empirical performance and prompt quality, with this value determined through ablation studies. We find that across subtasks, when we enforce $F_{\text{aux}}(p) \rightarrow [0, 100]$, that $\lambda = 0.25$ performs well empirically. This balanced approach ensures that the evolutionary process discovers prompts that are both effective and interpretable, while the LLM-based critique system provides a computationally efficient mechanism for maintaining semantic coherence throughout the optimization process. This novel modification reduces the need for more complicated mutation mechanisms adopted by other works, such as Promptbreeder.

\subsection{Evolutionarily Emergent Tool Synthesis}

\begin{figure}[H]
  \includegraphics[scale=0.4]{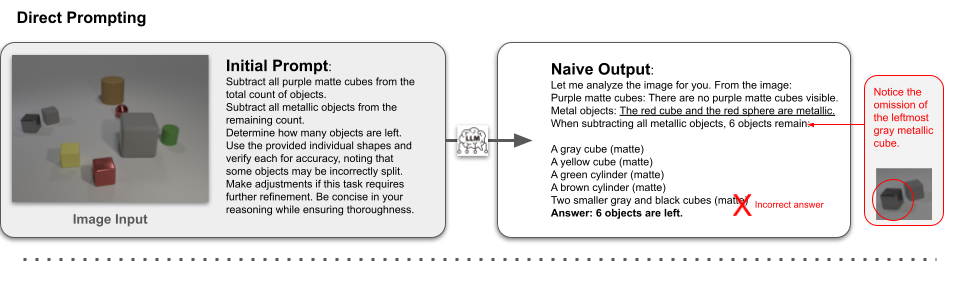}
  \caption{A naive example of directly prompting an input image using the original dataset prompt. Note that the LLM misses one of the metallic shapes, leading to an incorrect conclusion. Given the visual complexity of the input image, Vision Language Models may struggle to accurately analyze subcomponents without further guidance.}
  \label{fig:failure_ex}
\end{figure}

A key discovery in our evolutionary framework is the emergence of self-referential tool generation capabilities. Through our robust and performant evolutionary search procedure, as well as high-quality initial universes of mutation prompts and task prompts, we find that evolutionary search procedures for certain visual tasks yield task prompts that attempt to modify and re-ingest the input image(s) for multiple passes of reasoning. A successful example of this evolutionary reasoning is shown in Figure~\ref{fig:success_ex}, in contrast to an unsuccessful naive prompting example in Figure~\ref{fig:failure_ex}.

Rather than predefining a fixed tool universe, we observe that evolved system prompts naturally develop the ability to decompose problems into tool-like operations. We then leverage the natural capacity of LLMs to generate performant code from natural-language instructions by converting the natural language tool description into Python code with an auxillary LLM and executing it on the input image(s).

The evolutionary process operates on system prompts $s \in S$ that guide the primary language model $\mathcal{L}_1$ in processing inputs. Through mutation and selection pressure, these prompts evolve to incorporate structured tool suggestions enclosed in XML tags:

\begin{equation}
    \mathcal{L}_1(s, x) \rightarrow (...\texttt{<tool>}\tau_i\texttt{</tool>}...)_{i=1}^k
\end{equation}

where each $\tau_i$ represents a natural language description of a proposed tool operation. These tool suggestions emerge organically as the system discovers that breaking down complex tasks into composable operations improves performance. A secondary language model $\mathcal{L}_2$ acts as an interpreter, translating each tool suggestion into executable Python code:

\begin{equation}
    \mathcal{L}_2(\tau_i) \rightarrow c_i \in \mathcal{C}
\end{equation}

where $\mathcal{C}$ is the space of valid Python programs. This creates a flexible tool synthesis pipeline where $\mathcal{L}_2$ leverages its code generation capabilities to implement operations like image manipulation, mathematical computations, or data processing based on natural language descriptions.

The composed transformation on input $x$ becomes:

\begin{equation}
    T(x) = \text{eval}(c_k \circ ... \circ c_1)(x)
\end{equation}

where the composition emerges from the sequential application of synthesized tools. Critically, this approach allows for open-ended tool discovery, because the system isn't constrained by predefined tools. Additionally, this approach allows for recursive refinement, as the tool outpquts can be fed back into $\mathcal{L}_1$ for iterative processing. This represents a novel reasoning paradigm for traditional vision-language models, as they can have multiple iterative reasoning passes at the same image, allowing for more complex reasoning patterns such as examining different patches of the image multiple times, increasing the brightness/contrast of patches, and applying external models such as Meta's Segment Anything (SAM) tool \cite{meta-sam}. Due to the expressivity of natural language and the efficacy of LLMs in converting natural language instructions to executable code, tool usage patterns become increasingly effective on downstream tasks with respect to generation count. 

%This framework creates a powerful synergy: $\mathcal{L}_1$ evolves increasingly sophisticated strategies for breaking down problems, while $\mathcal{L}_2$ provides reliable implementation of the suggested operations. The evolutionary pressure drives the system toward discovering effective tool combinations, as prompts that generate useful and composable tool suggestions achieve higher fitness scores.%

The emergence of structured tool suggestions in evolved prompts indicates that the system has discovered a fundamental principle: complex tasks often benefit from decomposition into smaller, well-defined operations. Crucially, this discovery happens naturally through the evolutionary process, as prompts that effectively utilize this pattern tend to produce better results across diverse inputs, resulting in a iteratively-optimized final prompt.
yeah th

\section{Results}

Our experimental results demonstrate significant improvements across multiple vision-language reasoning benchmarks through evolutionary prompt optimization. In all cases we benchmark results on OpenAI's model 4o mini \citep{2023GPT4VisionSC}. Table~\ref{tab:results} presents a comprehensive comparison of our approach against several baselines, including the base model with no Chain-of-Thought prompting (4o mini), standard Chain-of-Thought prompting (+CoT) using \enquote{Let's think step by step} as in \citet{wei2023chainofthoughtpromptingelicitsreasoning}, and PromptBreeder (+PB). Our evolutionary tool synthesis approach (+Tools) achieves substantial gains across nearly all tasks, with particularly notable improvements in tasks requiring complex spatial and physical reasoning.\\

\begin{table}[ht]
    \centering
    \renewcommand{\arraystretch}{1.2}  % Increase row spacing
    \begin{tabular}{llrrrrr}
        \toprule
        \textbf{Benchmark} & \textbf{Task}              & \textbf{4o mini} & \textbf{+CoT} & \textbf{+PB} & \textbf{+Ours} & \textbf{++Tools} \\
        \midrule
        MathVista          & Visual QA                  & 49.5             & 51.0          & 49.6       & 53.3         & \textbf{60.5}   \\
                           & Figure QA                  & 58.6             & 60.1          & 58.7       & 61.5         & \textbf{64.1}   \\
                           & Math Word Problem          & 61.8             & 63.2          & 61.9       & 64.5         & \textbf{68.0}   \\
        \midrule
        M3CoT              & Geometry                   & 37.8             & 39.1          & 37.9       & 35.2         & \textbf{42.1}   \\
                           & Theory                     & 6.1              & \textbf{9.0}           & 6.2        & 6.3          & --              \\
                           & Physical Commonsense       & 42.6             & 43.9          & 42.7       & 47.2         & \textbf{61.7}   \\
        \midrule
        GeoBench-VLM       & Damaged Building Count     & 21.5            & 22.2         & 21.6      & 21.0        & \textbf{32.1}  \\
                           & Crop Type Classification   & 9.8            & \textbf{10.1}         & 9.9      & 9.8        & 10.0  \\
                           & Farm Pond Change Detection & 12.3            & 12.7         & 12.4      & 14.1        & \textbf{20.2}  \\
        \bottomrule
    \end{tabular}
    \caption{Performance across benchmarks showing the impact of different reasoning approaches. Best results for each task are bolded.\\[1ex]
    \footnotesize{CoT = Chain of Thought, PB = PromptBreeder, Ours = Our method (with vision initial population) without tool interpreter access, +Tools = our method with tool interpreter access. A dash (–) in the +Tools column indicates that Tools were not elicited due to the nature of the subtask, so the performance is identical to the Ours column for that subtask.}}
    \label{tab:results}
\end{table}

The experimental results clearly demonstrate that our evolutionary prompt optimization framework markedly enhances multimodal reasoning in vision-language models. As evidenced in Table~\ref{tab:results}, while the standard chain-of-thought (CoT) prompting yield only incremental improvements, our evolved prompts (denoted as “+Ours”) already push performance higher across tasks, and the addition of tool interpreter access (“++Tools”) consistently achieves the best outcomes. For example, in the MathVista benchmark, performance on Visual QA improves from 49.5 with 4o mini to 53.3 with our approach, and further to an impressive 60.5 when tool usage is enabled. Similar trends are observed across other tasks—including Figure QA and Math Word Problems—indicating that the evolutionary process not only refines prompt instructions for better task alignment but also facilitates the spontaneous emergence of sophisticated strategies such as hierarchical problem decomposition and dynamic tool synthesis. These findings underscore the potential of inference-time prompt evolution to unlock latent reasoning capabilities in vision-language models, thereby offering a scalable and efficient pathway toward more robust multimodal AI systems.

% \vspace{-10mm}
\subsection{Analysis of Emergent Behaviors}

A particularly interesting finding is the emergence of sophisticated tool-use patterns through evolution. The evolved prompts frequently develop structured approaches to problem decomposition, often breaking complex tasks into sequences of simpler operations. For instance, in the Math Word Problem task, we observe prompts that systematically partition large images into manageable sections.

\begin{figure}[htb]
    % \vspace{-5em}  % Reduce space above
    \setlength{\abovedisplayskip}{0pt}
    \setlength{\belowdisplayskip}{0pt}
    \centering
    \includegraphics[scale=0.5]{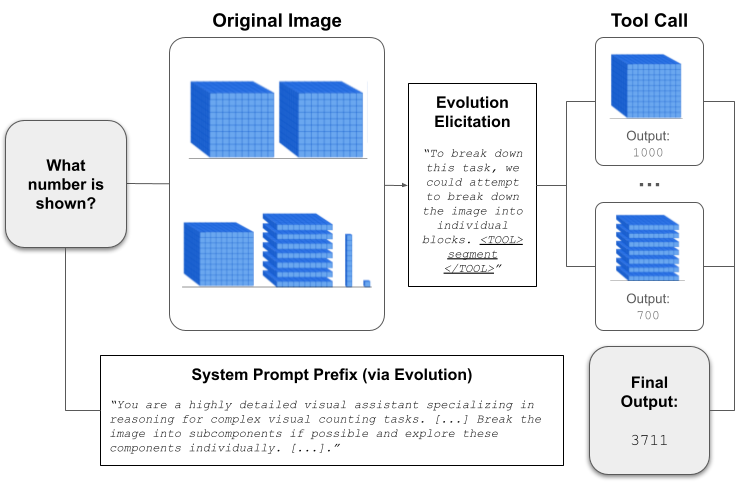}
    % \vspace{-5em}  % Reduce space below
\end{figure}

These behaviors emerged naturally through the evolutionary process, without explicit programming or human demonstration. Similarly, in physical reasoning tasks, the evolved prompts often exhibit a form of "mental simulation," breaking down complex physical scenarios into sequences of simpler state transitions.

The results demonstrate that our evolutionary framework not only improves raw performance metrics but also discovers interpretable and generalizable reasoning strategies. The emergent behaviors often mirror human problem-solving approaches, suggesting that the framework is finding natural and effective solutions to complex reasoning tasks.

\subsection{Generalization Experiments}
\noindent
\begin{minipage}[t]{0.55\textwidth}
    \vspace{0pt}%
    We also measure the generalization capabilities of our evolution framework compared to the base-line method in multimodal reasoning domains. Due to cost constraints induced by evaluating on larger datasets, we run these generalization experiments on MolmoE-1B-0924 \citep{deitke2024molmopixmoopenweights} rather than 4o mini. Evolutionary prompt optimization only poses significant utility on downstream tasks if our assumption holds--that the optimal system prompt generated through prompt optimization and evaluation on a train set generalizes to perform near-optimally on a withheld test set. We compare the baseline PromptBreeder \citep{fernando2023promptbreederselfreferentialselfimprovementprompt} method to our vision-language-specific approach (both with tool usage enabled and without), and find that both our approaches generate high-fitness system prompts with just 20-30\% of the total dataset, which is often under 20 individual samples. We attribute this to our improved LLM-augmented fitness function, which serves as an effective regularizer to the search process and yields higher sample efficiency. \end{minipage}%
\hfill
\begin{minipage}[t]{0.40\textwidth}
    \vspace{0pt}%
    \centering
    \includegraphics[width=\linewidth]{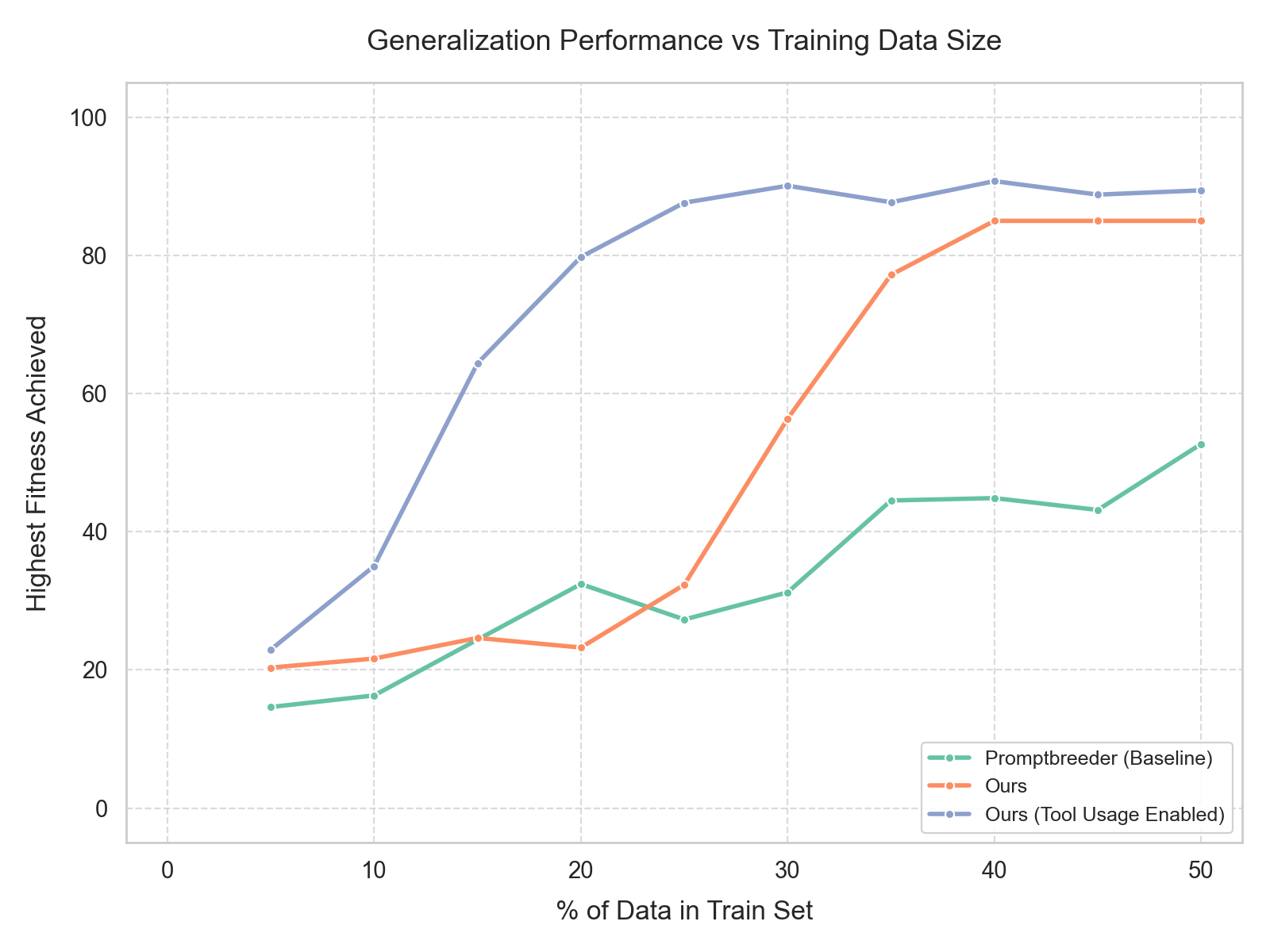}
    \captionof{figure}{Generalization performance of various prompt optimization techniques on Damaged Building Count vision-heavy reasoning task}
    \label{gen-chart}
\end{minipage}

\section{Discussion}

Our results demonstrate the remarkable potential of evolutionary prompt optimization for enhancing multimodal reasoning capabilities in vision-language models. This work offers several important insights about the nature of artificial reasoning, the power of guided search in prompt space, and the future of multimodal AI systems.

\subsection{Evolutionary Search as a Path to Advanced Reasoning}

Perhaps the most intriguing aspect of our findings is how evolutionary search, guided by LLM-based mutations, discovers sophisticated reasoning strategies without explicit programming or human demonstration. The framework's success across diverse tasks—from geometric reasoning to temporal change detection—suggests that the prompt space contains rich, generalizable programs for complex cognitive tasks. This emergence of advanced reasoning patterns through guided exploration has several important implications:

First, it demonstrates that LLMs can serve as effective mutation operators in evolutionary search, leveraging their understanding of language and task structure to make meaningful modifications to prompts. This is particularly noteworthy because it creates a form of "cognitive bootstrap," where the LLM's capabilities are used to evolve prompts that then enhance the same model's performance.

Second, the discovered prompts often exhibit surprisingly sophisticated behaviors, such as hierarchical decomposition, systematic scanning patterns, and implicit error checking. These strategies mirror human problem-solving approaches but emerge purely through the evolutionary process. This suggests that guided search in prompt space can uncover "natural" cognitive algorithms that are both effective and interpretable.

\subsection{System Prompts as Lightweight Neural Programs}

Our work underscores the importance of viewing system prompts as a powerful and underexplored form of neural programming. Unlike traditional neural architecture modifications or fine-tuning approaches, system prompts can encode complex computational strategies without changing model weights. We emphasize several advantages of this paradigm, including computational efficiency, architectural flexibility, and program-like expressivity:

First, our approach requires only inference-time computation for prompt evolution and no expensive model retraining. This makes it particularly attractive for deployment scenarios where computational resources are relatively limited at inference time.

Second, the evolved prompts can be added to existing vision-language models as a lightweight inference-time intervention, improving performance without drastic architectural changes. This modularity is valuable for practical applications, as its performance improves with performance-based optimizations for vision-language models.

Third, the prompts effectively serve as programs, capable of encoding sophisticated control flow, decomposition strategies, aggregation of subproblems, and error-handling mechanisms. This suggests that natural language itself can serve as a powerful programming medium for neural systems.

\subsection{The Case for Native Multimodal Reasoning}

The emergence and strong performance of image-based tool calling behavior within the evolved prompts motivates native multimodal reasoning as a future reasoning paradigm. Our method of decomposing complex multimodal tasks into smaller image-based patches, performing text-based reasoning on the subproblems, and then aggregating back up outperforms purely text-based reasoning methods across several subdomains. Through evolutionary tool usage, the system prompts learn to decompose complex tasks into smaller patches and reasoning over these patches. This can be seen as a primitive of reasoning natively over both text and image modalities flexibly, and hence motivates the development of more advanced multimodal reasoning models. Additionally, allowing the vision-language model to conduct multiple passes over the visual patches enables more robust feature extraction and relationship understanding. This iterative processing allows the model to build up a hierarchical representation of the visual information, where initial passes might capture low-level features and spatial relationships, while subsequent passes can integrate this information with higher-level semantic understanding and task-specific goals. The emergence of this multi-pass behavior in our evolved prompts suggests that effective multimodal reasoning requires not just the ability to process different modalities, but also the capability to dynamically revisit and reinterpret information as the reasoning process unfolds. This finding has implications for architectural design choices in future multimodal systems, particularly in how attention mechanisms and information flow are structured across multiple reasoning steps.

\subsection{Scaling Guided Search towards Omnimodal AI}

Our findings have several implications for the future development of vision-language model systems focused on reasoning.

Guided Search as a Development Paradigm: The success of our evolutionary framework suggests that guided search in prompt space could be a powerful paradigm for developing AI capabilities more broadly. This approach might be particularly valuable for discovering reasoning strategies in domains where human intuition is limited \citep{deepseekai2025deepseekr1incentivizingreasoningcapability}.

Towards Omnimodal AI: The framework's ability to evolve effective multimodal reasoning strategies motivates a path toward truly omnimodal AI systems that can seamlessly integrate information across different modalities. This could be particularly important for real-world applications that require understanding complex multi-modal scenarios.

Novel Test-Time Scaling Laws: Our approach achieves notable emergence through continual system prompt evolution, with emergent paradigms such as multimodal tool usage and dynamic programming emerging as we increase the amount of compute expended on evolution. This suggests a potential future inference-time scaling law in prompt / system program space, which may act as complementary to the recent advances in inference-time compute.

\subsection{Limitations and Future Work}

Despite the promising performance of our evolutionary seach procedure for discovering multimodal reasoning paradigms, several limitations warrant further investigation. First, the failure to improve performance on theoretical tasks suggests limitations in the current framework's ability to handle highly abstract reasoning. Our evaluations are currently confined to a limited set of benchmarks (MathVista, M3CoT, GeoBench-VLM) and a single model (4o mini), underscoring the need for broader testing across diverse tasks and architectures to establish generalizability. Additionally, our reliance on the underlying LLM for both mutation and fitness evaluation introduces variability in prompt quality and tool synthesis reliability, highlighting the necessity for enhanced error-handling and verification mechanisms, potentially through ensemble approaches. The extra inference-time computational overhead, while lower than model retraining, could still be pose challenges for deployment in resource-constrained or latency-sensitive environments, indicating that more efficient, parallelizable evolution strategies are needed. Exploring other search paradigms for prompt optimization on multimodal reasoning tasks, such as reinforcement learing or reasoning in latent spaces, are promising and well-motivated given our positive results. Finally, while our framework naturally uncovers sophisticated behaviors such as hierarchical decomposition and emergent tool synthesis, the mechanisms driving these phenomena remain underexplored, and integrating our few-shot, inference-time optimization with continuous learning paradigms presents an exciting yet open challenge. 

\bibliography{iclr2025_conference}
\bibliographystyle{iclr2025_conference}

\newpage
\appendix
\section{Appendix}
\subsection{Full Evolutionary Algorithm Outline}
\begin{algorithm}[h]
\caption{Evolutionary Prompt Optimization}
\label{alg:evolutionary_prompt_optimization}
\begin{algorithmic}[1]
    \State \textbf{Input:} 
        \begin{itemize}
            \item Training set $A$
            \item Initial prompt population $\mathcal{P}_0$ of size $N$
            \item Mutation prompt space $\mathcal{M}$ and hyper-mutation space $\mathcal{H}$
            \item Fitness function $F(p) = (1-\lambda)F_{\text{task}}(p) + \lambda F_{\text{aux}}(p)$
            \item Maximum number of generations $G$
        \end{itemize}
    \State \textbf{Output:} Optimized prompt $p^*$

    \State \textbf{Initialization:}
    \State Set generation counter $g \gets 0$
    \State Initialize prompt population $\mathcal{P}_0$ with $N$ candidate prompts
    \While{$g < G$}
        \State \textbf{Selection and Evaluation:}
        \State Randomly sample two distinct prompts $p_1, p_2 \in \mathcal{P}_g$
        \State Compute fitness scores $F(p_1)$ and $F(p_2)$
        \If{$F(p_1) \geq F(p_2)$}
            \State Set winner $p_w \gets p_1$ and loser $p_l \gets p_2$
        \Else
            \State Set winner $p_w \gets p_2$ and loser $p_l \gets p_1$
        \EndIf
        \State \textbf{Mutation:}
        \State Sample a mutation prompt $m \in \mathcal{M}$
        \State Generate mutated prompt: $p'_w \gets \mathcal{L}(m \oplus p_w)$
        \State \textbf{Population Update:}
        \State Update population: $\mathcal{P}_{g+1} \gets (\mathcal{P}_g \setminus \{p_l\}) \cup \{p'_w\}$
        Apply hyper-mutation operators using $h \in \mathcal{H}$.
        \State Increment generation: $g \gets g + 1$
    \EndWhile
    \State \textbf{Return:} $p^* \gets \argmax_{p \in \mathcal{P}_G} F(p)$
\end{algorithmic}
\end{algorithm}

\subsection{Naive Prompting Example - Failure}

\clearpage
\Needspace{22cm}
\subsection{Evolutionary Prompting Example - Success}
\begin{figure}[H]
  \centering
  \includegraphics[scale=0.4]{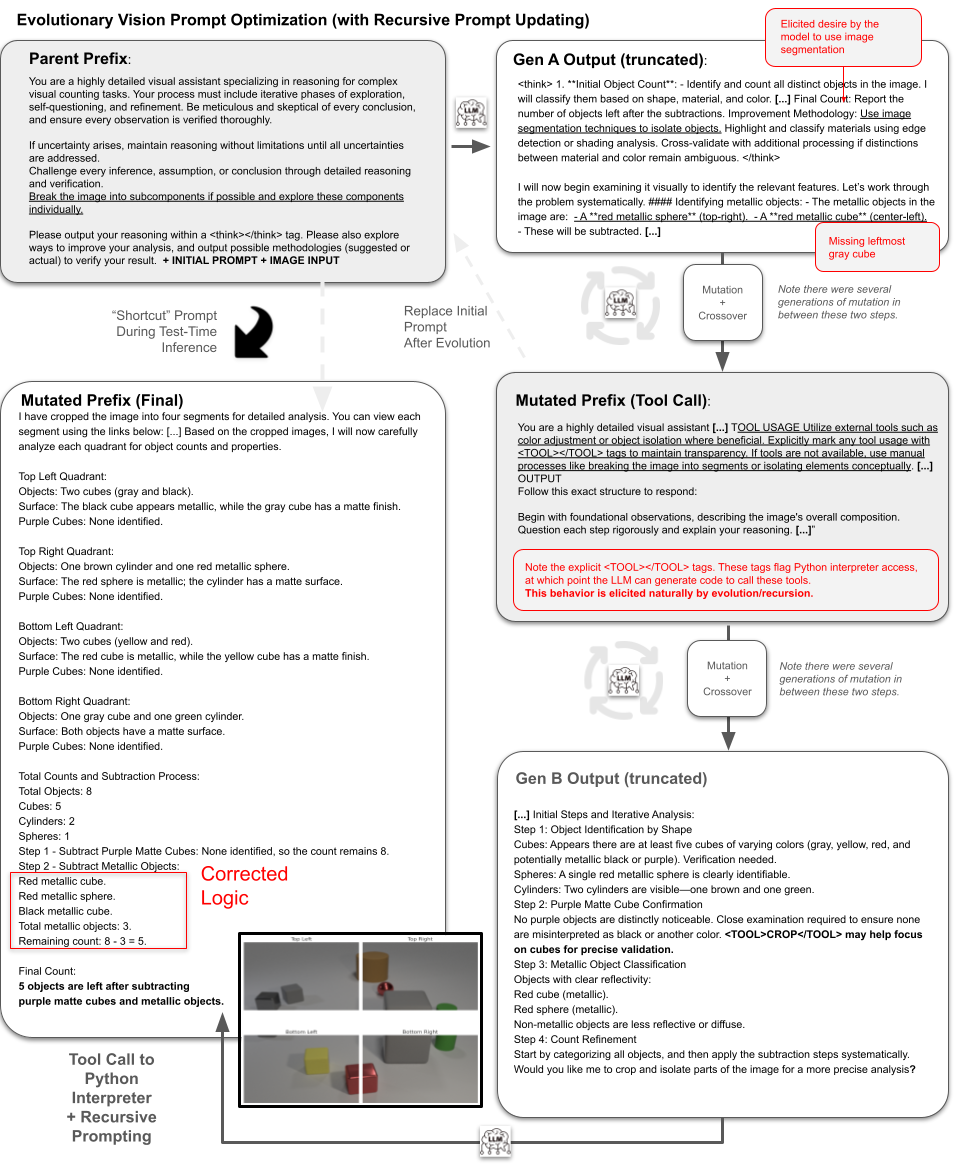}
  \caption{Walk through of an example where an initial prompt fails to elicit a correct answer, while a successful evolutionarily optimized prompt including a tool call (cropping) succeeds. Via the evolved prompt, the model elicits a tool call that crops the original image, allowing the LLM to better ingest the image's contents. With the improved quadrant division of visual analysis, the model is able to correctly answer the question.}
  \label{fig:success_ex}
\end{figure}

\begin{figure}[h]
    \subsection{Prompt Fitness as a function of Evolution Time}
    \centering
    % First image
    \begin{minipage}{\textwidth}
        \centering
        \includegraphics[width=1.0\textwidth]{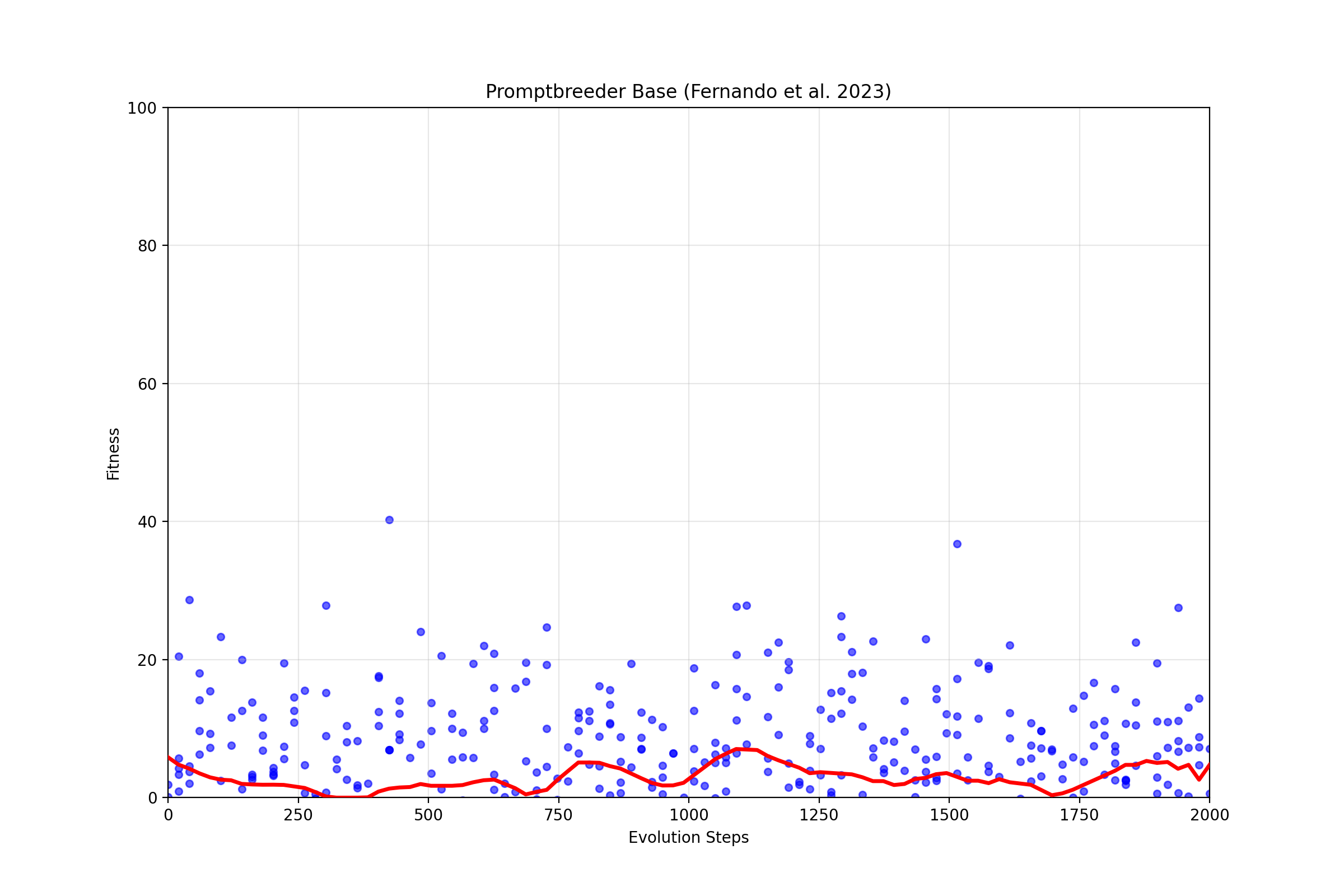}
        \caption{Baseline evolutionary prompt optimization method (Promptbreeder, Fernando et. al. 2023) fails to generalize to vision-language reasoning domains. We find this is because their instruction-following prompting for LLM mutation, hypermutation, and their initial universes are not suited for vision-language reasoning tasks. }
    \end{minipage}
    % Second image
    \begin{minipage}{\textwidth}
        \centering
        \includegraphics[width=1.0\textwidth]{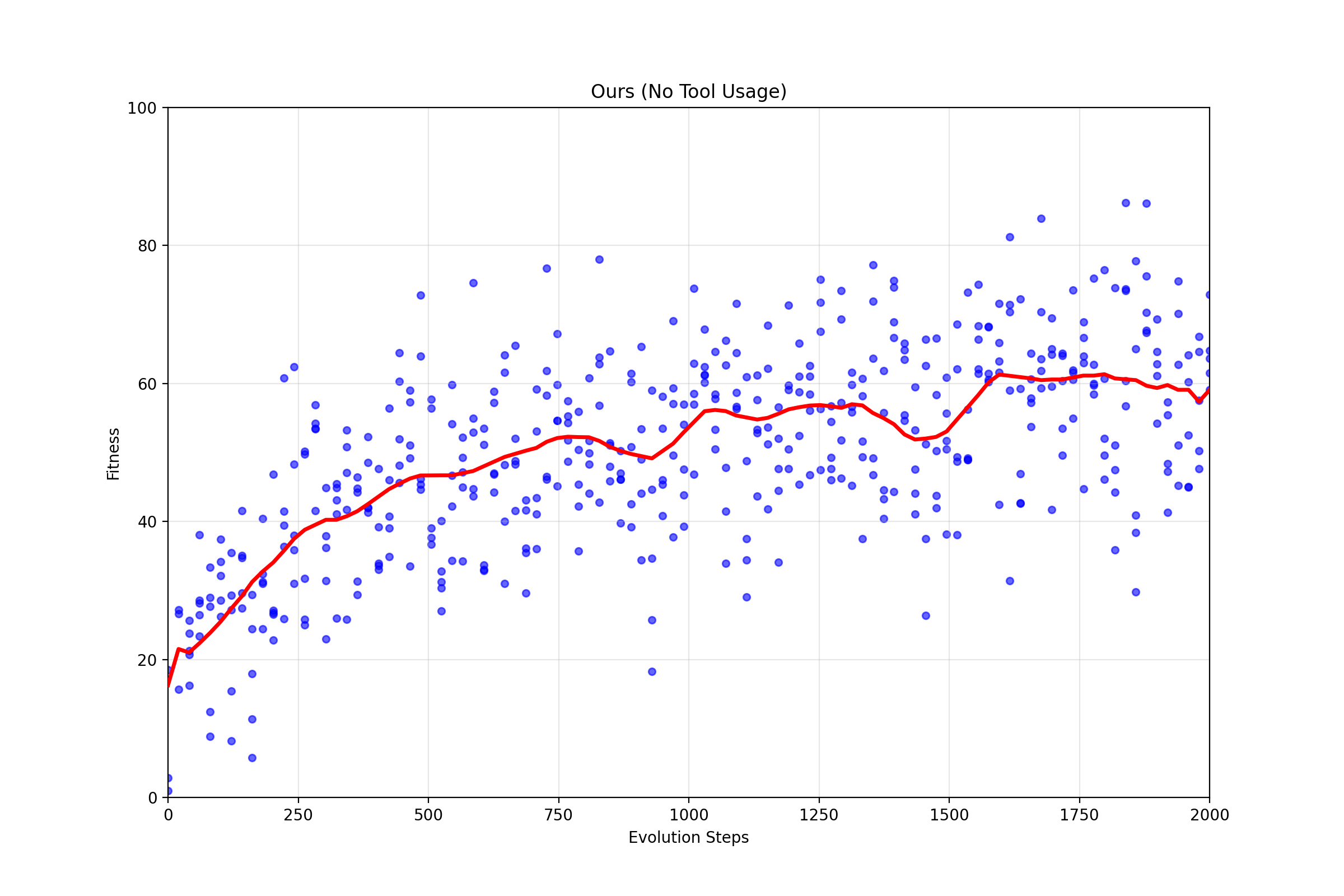}
        \caption{Our naive method outperforms baselines in the evolution process, due to significant improvements in mutation methods, our auxillary loss preventing significant and nonsensical deviations from current task prompts, and our initial universes of task prompts, mutation prompts, and hypermutation prompts, that are tuned specifically for image tasks. }
    \end{minipage}
    \label{fig:success_ex}
\end{figure}
\clearpage
\begin{figure}[H]
    % Third image
    \begin{minipage}{\textwidth}
        \centering
        \includegraphics[width=1.0\textwidth]{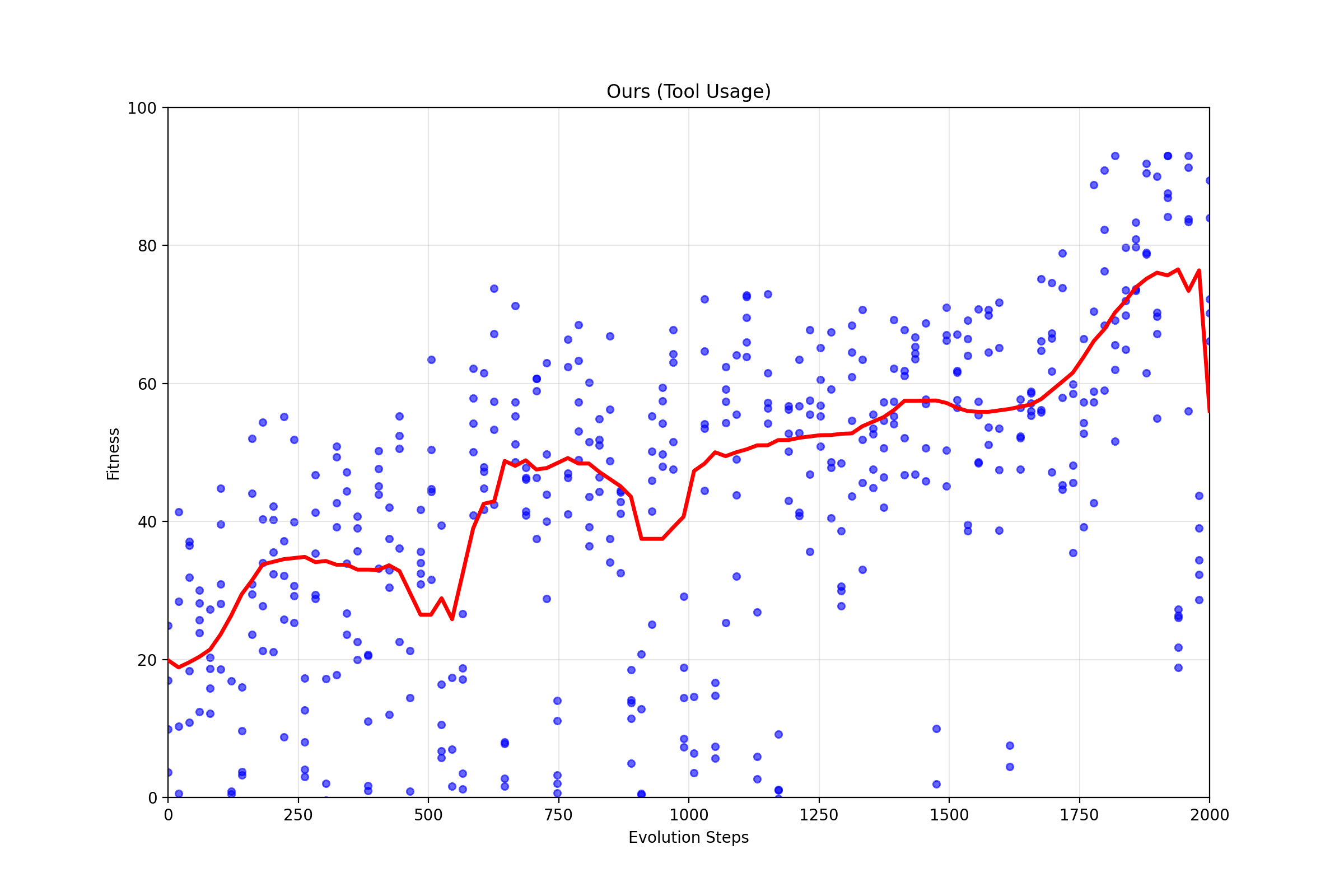}
        \caption{With tool use enabled, and an improved set of mutation and hypermutation prompts that encourages emergence of tool use, population fitness scales positively with time. The notable drops in average performance (red curve) are the critical windows in which tool usage emerges through evolution. Initially, fitness falls, because the tool usage and reasoning paradigms are nascent, but as they are evolved more, they become high-performing. Towards the end, the evolution process guides the highest-performing system prompts towards another layer of tool calling. This emergent strategy results in an immediate drop in performance, due to its nascence and incompleteness. We hypothesize that continuing the evolution process further would lead to even more advanced reasoning paradigms like these results indicate, but under our fixed computation budget, this remains an avenue for future research. }
    \end{minipage}
    \label{fig:success_ex_tools}
\end{figure}
\subsection{Auxiliary LLM Critic Implementation Details}
The auxiliary critic component ensures evolved prompts maintain coherence and stay aligned with task objectives. Implemented using GPT-4o-mini, the critic evaluates prompts through a multi-dimensional rubric designed to prioritize clarity, logical structure, and task relevance. We assign weights to different components of the given task prompt's quality, which are determined empirically. The critic operates via a structured evaluation template that emphasizes task fidelity:
\begin{verbatim}
Evaluate this prompt for:

Relevance to the stated visual reasoning task

Logical flow between instructions

Clarity of language and specificity
Score each dimension 1-5, then compute weighted total (0-100).
Flag any instructions that deviate from the task's core requirements.
\end{verbatim}
\clearpage
\subsection{Mutation Prompts and Task Prompts Initial Universes}

\begin{table}[h]
\centering
\caption{Sample Vision-Language Starting Prompts}
\label{tab:evolved-prompts}
\renewcommand{\arraystretch}{1.18}
\begin{tabular}{@{}p{9.2cm} p{5.5cm}@{}}
\toprule
\textbf{Initially Evolved Prompt (Vision-Task Aligned)} & \textbf{Reasoning Strategy} \\
\midrule

\textit{“Generate a diagram highlighting the fundamental shapes and key objects in the image. Use these as anchors to guide your final answer (such as a numeric value).”}
& 
Focuses on style-invariant structures to capture essential spatial and content information. \\[6pt]

\textit{“Translate the relevant visual features into symbolic or textual notations, aiming for both clarity and accuracy. Then refine this representation to produce a final answer.”}
& 
Balances interpretability and precision, handling trade-offs between simplicity and completeness. \\[6pt]

\textit{“Segment the image according to the task requirements, focusing on regions most relevant to the question. Prioritize these segments for deeper analysis.”}
& 
Applies task-specific segmentation to isolate key areas, reducing distraction from less important regions. \\[6pt]

\textit{“Iteratively refine your approach by generating bounding boxes or region proposals for the image. Retain only the proposals that significantly improve the clarity or correctness of your final result.”}
& 
Uses a population-based or iterative mechanism to refine localized views of the image. \\[6pt]

\textit{“Simulate common edge cases or distortions (like occlusion and unusual lighting) to see how they affect your answer. Refine the prompt steps that cause ambiguous or incorrect responses.”}
& 
Incorporates robustness testing and iterative prompt fixes for improved fault tolerance. \\[6pt]

\textit{“Construct a hierarchical representation of objects in the image, capturing relationships at multiple scales. Merge the partial findings into one cohesive conclusion.”}
& 
Organizes local and global features in a multi-scale structure for more holistic reasoning. \\[6pt]

\textit{“Generate several possible answers for the question by varying the approach. Compare how well each aligns with the visual details, and select the most fitting explanation.”}
& 
Uses contrastive evaluation to identify the answer best supported by the evidence in the image. \\[6pt]

\textit{“Apply a mix of symbolic and sub-symbolic steps to interpret the image. For instance, if the question involves counting objects, express it in conditional form (IF more than X, THEN...). Evaluate which approach yields the clearest final result.”}
& 
Combines rule-based reasoning with learned representations for interpretable and flexible analysis. \\[6pt]
\bottomrule
\end{tabular}
\end{table}

\clearpage
\makeatletter
\setlength{\@fptop}{0pt} % forces floats on a float page to stick to the top
\makeatother

\begin{table}[t]
\centering
\caption{Sample Mutator Prompts}
\label{tab:sample-mutator-prompts}
\renewcommand{\arraystretch}{1.2}
\begin{tabular}{p{11.5cm}}
\toprule
\textbf{Prompt} \\
\midrule
\textit{“Rewrite the instruction so that it focuses on \textbf{breaking down} any complex parts into simpler steps. Include a helpful tip for someone struggling.”} \\[6pt]

\textit{“Note there is likely a \textbf{critical error} in the last response. Please revise and share reasoning akin to a careful human. A corrected version would be:”} \\[6pt]

\textit{“Rephrase the instruction as if you are \textbf{guiding someone who does not have visual stimulus}, making sure every detail is crystal clear.”} \\[6pt]

\textit{“Imagine you must \textbf{teach this instruction to a peer} who could be easily confused. Simplify it, but offer one surprising or creative example.”} \\[6pt]

\textit{“Imagine a \textbf{shortcut} for this task, if you had infinite resources and capabilities. SHORTCUT=”} \\[6pt]

\textit{“Flip the point of view: rewrite the instruction as if \textbf{the user is already an expert}, and you are simply double-checking their approach.”} \\[6pt]

\textit{“Break the instruction into \textbf{two different methods}—one for someone who learns best by doing, and another for someone who prefers planning.”} \\[6pt]

\textit{“If useful for the task, explicitly highlight additional external functionality that you would find beneficial using \textbf{XML tags}.”} \\[6pt]

\textit{“Create a more \textbf{visual-oriented version} of the instruction by prompting the user to sketch out key steps or components before proceeding.”} \\[6pt]

\textit{“Rewrite the instruction in a \textbf{step-by-step checklist} format, then add a final insight or reminder that ensures the goal is met.”} \\

\bottomrule
\end{tabular}
\end{table}

\begin{table}[h]
\centering
\caption{Sample System Prompt via Evolution}
\label{tab:concise-prompt}
\renewcommand{\arraystretch}{1.18}
\begin{tabular}{@{}p{15cm}@{}}
\toprule
\textbf{Prompt} \\
\midrule
You are an expert visual reasoning assistant specializing in detailed image analysis and precise counting. Break the image into key components and analyze each individually using iterative reasoning. Document your steps briefly and employ image tools (e.g., cropping, segmentation) when beneficial. Clearly mark any tool usage with \texttt{<TOOL>} tags (e.g., \texttt{<TOOL>CROP</TOOL>}, \texttt{<TOOL>Segment</TOOL>}); if no tools are used, output \texttt{<TOOL>n/a</TOOL>}. Begin with an overview of the image, provide concise reasoning and segmentation strategies, and conclude with your final verified result. \\
\bottomrule
\end{tabular}
\end{table}

\end{document}